\documentclass[10pt,twocolumn,letterpaper]{article}
\usepackage{spconf}
\usepackage[linesnumbered,ruled,vlined]{algorithm2e}

\SetKwComment{Comment}{$\triangleright$\ }{}
\usepackage{times}
\usepackage{epsfig}
\usepackage{cite}
\usepackage{graphicx}
\usepackage{amsmath}
\usepackage{amssymb}
\newcommand{\R}{{\mathbb R}}
\usepackage{multirow}
\usepackage{caption}
\usepackage{adjustbox}

\usepackage[breaklinks=true,bookmarks=false]{hyperref}

\title{Survey of Image Based Graph Neural Networks}
\name{U. Nazir$^{\star \dagger}$ \qquad H. Wang$^{\star \dagger}$ \qquad M. Taj$^{\star}$ }
\address{$^{\star \dagger}$ \{scun, H.E.Wang\}@leeds.ac.uk \\
$^{\star}$\{murtaza.taj\}@lums.edu.pk }

\begin{document}
\ninept


\maketitle

\begin{abstract}
In this survey paper, we analyze image based graph neural networks and propose a three-step classification approach. We first convert the image into superpixels using the Quickshift algorithm so as to reduce $30\%$ of the input data. The superpixels are subsequently used to generate a region adjacency graph. Finally, the graph is passed through a state-of-art graph convolutional neural network to get classification scores. We also analyze the spatial and spectral convolution filtering techniques in graph neural networks. Spectral-based models perform better than spatial-based models and classical CNN with lesser compute cost. 
\end{abstract}


\section{Introduction}
Deep learning, particularly convolutional neural networks have in the recent past revolutionized many machine learning tasks. Examples include image classification, video processing, speech recognition, and natural language processing. These applications are usually characterized by data drawn from the Euclidean space. Recently, many studies on extending deep learning approaches for graph data have emerged~\cite{henaff2015deep,defferrard2016convolutional,jain2016structural,kipf2016semi,wang2018deep,satorras2018few,narasimhan2018out,hu2018relation,gu2018learning, wang2018zero,lee2018multi,qi2018learning,marino2016more,kampffmeyer2019rethinking, edwards2016graph, liu2020cnn, fey2019fast, wan2019multiscale, qi2018stagnet, zhou2019relation, zhang2020spatio, tompson2014joint}. The motivation for these studies stems from the emergence of applications in which data is drawn from non-euclidean domains and then represented as graphs so as to capture the complex relationships and inter-dependency between objects. Indeed, many datasets and associated problems can be more naturally represented and analyzed as graphs~\cite{bliss2013confronting}. For instance, graph neural networks (GNNs) have been increasingly used for applications such as molecule and social network classification \cite{knyazev2018amer} and generation \cite{simonovsky2017dynamic}, 3D Mesh classification and correspondence \cite{fey2018splinecnn}, modeling behavior of dynamic interacting objects \cite{kipf2018neural}, program synthesis \cite{allamanis2017learning}, reinforcement learning tasks \cite{bapst2019structured} and many other exciting problems.

\begin{figure}[t]
\centering
\scalebox{1}{
	\begin{tabular}{ccc}
        \includegraphics[width=.28\columnwidth]{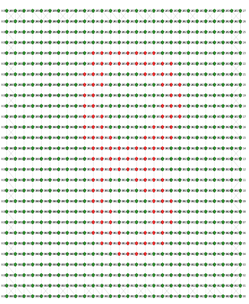} &
        \includegraphics[width=.28\columnwidth]{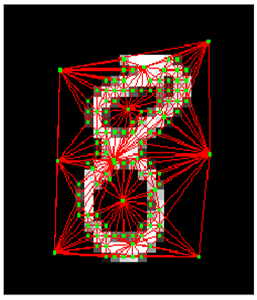} &
        \includegraphics[width=.28\columnwidth]{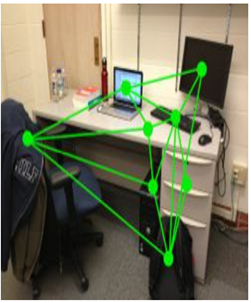} \\
        (a) & (b) & (c) \\
		\end{tabular}}
	\caption{An illustration of (a) Pixel-based graph, (b) Superpixel-based graph (c) Object-based graph~\cite{jiang2013hallucinated}.}
	\label{fig:gcat}
\end{figure}

While the utility of graph neural networks for emerging applications is promising, the complexity of graph data imposes significant challenges on many existing machine learning algorithms. For instance, in the area of image processing, the use of Graph Convolutional Network (GCN) is still limited to a few examples only~\cite{kampffmeyer2019rethinking,wang2018zero, lee2018multi}. By some carefully hand-crafted graph construction methods or other supervised approaches, images can be converted to structured graphs capable of processing by GCNs. In these GNNs, each pixel of an image is considered as a graph node~\cite{edwards2016graph} which is cumbersome and in many cases unnecessary. Instead of learning from individual image pixels, the use of 'superpixels' addresses this concern \cite{liang2016semantic, knyazev2019image} and helps in reducing the graph size and thereby the computational complexity. Graphs also allow us to impose a relational inductive bias in data for example via prior knowledge. For instance, in case of human pose, the relational bias can be a graph of skeleton joints of a human body \cite{yan2018spatial}. Similarly, in case of videos, relational bias can be a graph of moving bounding boxes~\cite{wang2018zero, marino2016more}. Another example is making inferences about facial attributes and identify by representing facial landmarks as a graph \cite{antonakos2015active}. The literature on application of GNNs on images can be broadly classified in to three groups (a) pixel-based graphs, (b) superpixel-based graphs and (c) object-based graphs - sample illustrations of these three methods are shown in Fig. \ref{fig:gcat}. In addition to providing a comprehensive review of graph techniques for images' superpixels, this paper paper also makes notable contribution by introducing new  taxonomy based on how graph represents an image as summarized in Table~\ref{tab:proposals}.

\begin{table}[b]
    \caption{Classification of Proposals}
    \centering
    \begin{tabular}{cc}
    \hline
    Graph Type   & Proposals \\ \hline
    Pixel-based Graph & \cite{defferrard2016convolutional, edwards2016graph, liu2020cnn} \\
    Superpixel-based Graph & \cite{liang2016semantic, knyazev2019image, fey2019fast, liu2020cnn, wan2019multiscale}\\
    Object-based Graph & \cite{jain2016structural,qi2018stagnet, zhou2019relation, zhang2020spatio, jiang2013hallucinated, tompson2014joint}\\
    \hline
    \end{tabular}
     \label{tab:proposals}
\end{table}

\section{Image Based GNNs}
GNNs on images are characterized by unique challenges with respect to their implementation. Most of the graph neural frameworks~ \cite{defferrard2016convolutional, edwards2016graph, liu2020cnn} are designed for dense representations such as pixel-based graphs. One can achieve effective parallelization on irregular sparse graphs (e.g., superpixel-based graphs) by using fast localized spectral filtering~\cite{defferrard2016convolutional} or hierarchichal multi graph spatial convolutional neural networks \cite{knyazev2019image}. Pixel, superpixel and object-based graphs have been extensively used in the literature as summarized in Table \ref{tab:proposals}. For subsequent processing, superpixels have been widely used as an effective way to reduce the number of image primitives. The literature includes numerous methods for determining a superpixel representation from an image, each with different strengths and weaknesses. Recently, many DNN-based methods to identify superpixels have been proposed~\cite{yang2020superpixel,jampani2018superpixel}. But the most popular of practices in GNN literature (on account of generally good results and low compute complexity) are SLIC~\cite{achanta2012slic}, Quickshift~\cite{vedaldi2008quick} and Felzenszwalb~\cite{felzenszwalb2004efficient}. Details of these methods are presented in the following.
\begin{figure}[t]
\centering
\scalebox{1}{
	\begin{tabular}{ccc}
	    \includegraphics[width=.28\columnwidth]{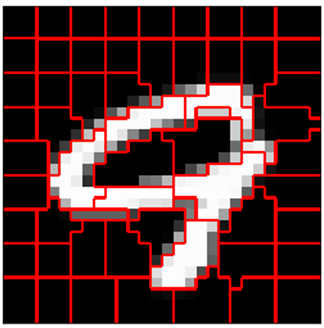} &
	    \includegraphics[width=.28\columnwidth]{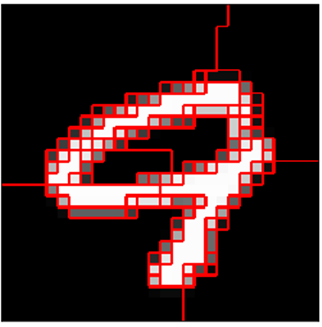} &
        \includegraphics[width=.28\columnwidth]{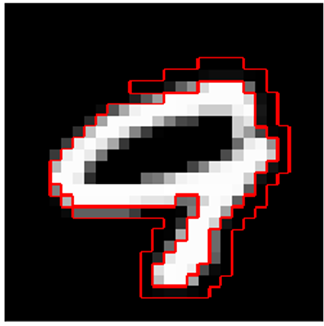}
        \\
        \includegraphics[width=.28\columnwidth]{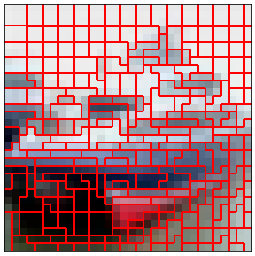} &
	    \includegraphics[width=.28\columnwidth]{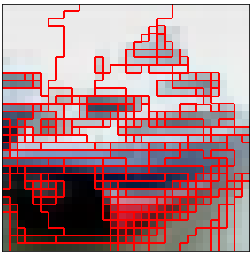} &
        \includegraphics[width=.28\columnwidth]{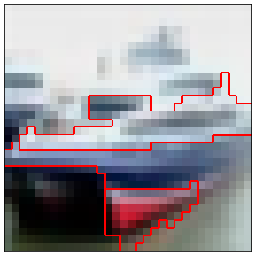}
        \\
        SLIC & Quickshift & Felzenszwalb  \\
		\end{tabular}}
	\caption{Superpixel segmentation techniques on MNIST digit: 9 and CIFAR-10 image: Ship.}
	\label{fig:superpixels}
\end{figure}
\subsection{Superpixel segmentation techniques}
\subsubsection{SLIC}
The SLIC (simple linear iterative clustering)~\cite{achanta2012slic} algorithm simply performs iterative clustering approach in the 5D space of color information and image location. The algorithm quickly gained momentum and is now widely used due to its speed, storage efficiency, and successful segmentation in terms of color boundaries. However, the limitation of SLIC is that it often captures the background pixels as shown in Fig. ~\ref{fig:superpixels} -- Column 1, and therefore does not significantly help in data reduction for graph generation.
\subsubsection{Quickshift}\label{qs}
Quickshift~\cite{vedaldi2008quick} is a relatively recent 2D algorithm that is based on an approximation of kernelized mean-shift~\cite{comaniciu2002mean}. It segments an image based on the three parameters: $\epsilon$ for the standard deviation of the Gaussian function, $\alpha$ for the weighting the color term and $S$ to limit the calculating a window size of $S \times S$. Therefore, it belongs to the family of local mode-seeking algorithms and is applied to the 5D space consisting of color information and image location. One of the benefits of Quickshift is that it actually computes a hierarchical segmentation on multiple scales simultaneously. As shown in Fig.~\ref{fig:superpixels} -- Column 2, it does not capture background pixels and also reduces $30\%$ of input data for graph generation. 
\subsubsection{Felzenszwalb}
This fast 2D image segmentation algorithm, proposed in ~\cite{felzenszwalb2004efficient}, has a single scale parameter that influences the segment size. The actual size and number of segments can vary greatly, depending on local contrast. This segmentation appeared to be less suitable in tests on a series of images, as its parameters requires a special adjustment and consequently a static choice of this parameter leads to unusable results. As shown in Fig.~\ref{fig:superpixels} -- Column 3, it only captures the pixels corresponding to the region of interest pixels but performs poorly in graph generation procedure.
\begin{figure}[h]
\centering
\scalebox{1}{
	\begin{tabular}{ccc}
        \includegraphics[width=.28\columnwidth]{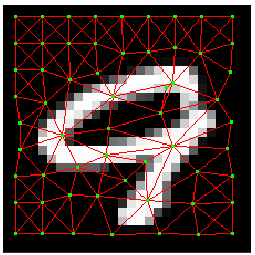} &
	    \includegraphics[width=.28\columnwidth]{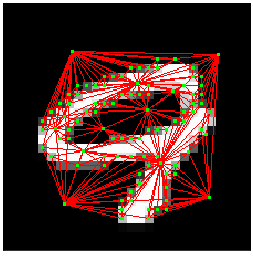} &
        \includegraphics[width=.28\columnwidth]{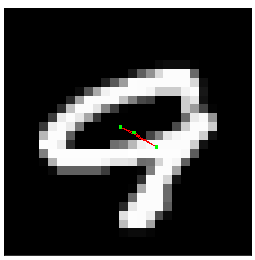}\\
        \includegraphics[width=.28\columnwidth]{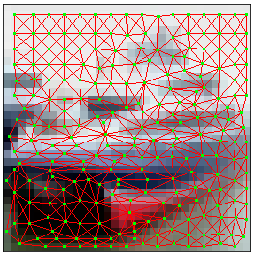} &
	    \includegraphics[width=.28\columnwidth]{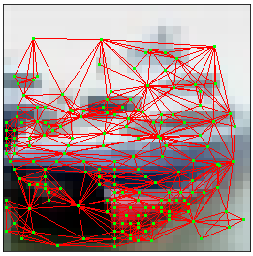} &
        \includegraphics[width=.28\columnwidth]{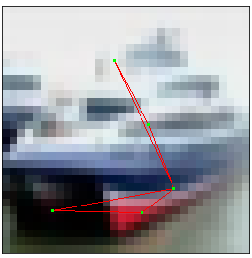}
        \\
		\end{tabular}}
	\caption{Region Adjacency Graphs (RAG) generation from SLIC, Quickshift and Felzenszwalb superpixels respectively.}
	\label{fig:RAGs}
\end{figure}
\subsection{Graph generation from superpixels}
This subsection discusses two methods used for generating graphs from superpixels. These are presented in the following.
\subsubsection{Region Adjacency Graph (RAG)}\label{rag}
After using a superpixel segmentation technique, a Region Adjacency Graph (RAG) is generated by treating each superpixel as a node and adding edges between all directly adjacent superpixels. Note that this differs from \cite{monti2017geometric} as it only connects to adjacent neighbors as compared to KNN. Each graph node can have associated features, providing an aggregate information based on characteristics of the superpixel itself. The superpixel segmentation technique using Quickshift is one way to divide the image into homogeneous regions. The regions obtained in the segmentation stage are represented as vertices and relations between neighboring regions are represented as edges. The search for the most similar pair of regions is repeated several times per iteration and every search requires $\mathcal{O}(N)$ region similarity computations. The graph is utilized so that the search is limited only to the regions that are directly connected by the graph structure. As we can see in Fig.~\ref{fig:RAGs}, the representation of an image via a graph $G$, which consists of superpixels representation $S$, has far fewer nodes in contrast to the grid representation of the original image. For graph generation, Quickshift (see Fig.~\ref{fig:RAGs} -- Column 2) performs well as compared to SLIC and Felzenszwalb.
\subsubsection{K-Nearest-Neighbors Graph (KNNG)}
In a KNN graph, each vertex chooses exactly $K$ nearest neighbors to connect. Superpixel graphs have connections that span more than one neighbour level, with edges formed with the $K$ nearest neighbours~\cite{monti2017geometric}. For more details on graph generation, see \cite{monti2017geometric}. For subsequent processing, RAGs have less compute complexity as compared to KNNG. This is the reason for passing RAGs to a graph CNN so as to obtain good classification accuracy. 
\begin{table}[h]
	\caption{Notations for Algorithms}
	\label{notations}
	\renewcommand{\arraystretch}{2.5}
	\centering
	\begin{adjustbox}{width=0.51\textwidth}
	\begin{tabular}{cc|cc}
		\hline
		Variable & Description & Variable & Description \\ \hline
		$\mathcal{E}$ & Set of edges &
		$\mathcal{V}$& Set of vertices \\
		$n$ & $|\mathcal{E}|$ &  
		$m$ & $|\mathcal{V}|$ \\
		$\mathcal{N}(v)$& Set of neighbors of vertex $v$ &
	    ${D_G}$ & Diagonal matrix of associated graph (G) \\ 
		$M_G$ & Region adjacency matrix &
		$W_G$ & Weighted region adjacency matrix \\ 
		$L_G$ & Laplacian Matrix &
		 $\{\lambda_l\}_{l=0}^{n-1}$ & Frequencies of graph \\ 
		$ U $ &  Fourier basis & 
		$\hat{x}$ & The graph Fourier transform of signal $x$ \\ 
		$h_\theta(\Lambda_G)$&  Non-parametric filter &
		$L$ & Number of GNN layers \\
		$h_v$& Input feature vector of vertex $v$ &
		$h_v^{l}$ & Hidden feature vector of vertex $v$ \\
		$h_v^{L}$ & Output feature vector of vertex $v$ &
		$\phi_v^{l}$& Node and edge combination feature of layer l \\
		\hline
	\end{tabular}
	\end{adjustbox}
\end{table}
\begin{table*}
    \caption{Evaluation Results showing comparison between classical CNN as well as state-of-the-art GCNs. }
    \label{tab:eval}
    \centering
    \begin{adjustbox}{width=0.95\textwidth}
    \begin{tabular}{cccccc}
    \hline
    Input   & CNN/GCNN Type & Model  &  Datasets & Accuracy (\%) & Compute Cost \\ \hline
    \multirow{2}{*}{Pixel based RAG}& \multirow{2}{*}{Spectral Convolution Filtering} & ChebNet~\cite{defferrard2016convolutional} & MNIST & 99.14  & $\mathcal{O}(\log (K |\mathcal{E}|))$\\
     &  & ChebNet~\cite{defferrard2016convolutional}  & CIFAR-10 & 98 & $\mathcal{O}(\log(K |E|))$\\ \hline
    \multirow{4}{*}{Superpixel based RAG} &  \multirow{4}{*}{Spatial Graph Filtering} & EGCNN ~\cite{fey2019fast} & MNIST & 98.025  & $\mathcal{O}(\log( |\mathcal{V}^2|))$\\
    & & GCN baseline & MNIST & 86.358 & $\mathcal{O}(\log( Kmh_i^l + Kn{h_i^l}^2))$\\
    & & EGCNN ~\cite{fey2019fast}&  CIFAR-10 & 75.230 & $\mathcal{O}(\log (\mathcal{V}^2\Large))$ \\
    & & GCN baseline & CIFAR-10 & 53.606 & $\mathcal{O}(\log( Kmh_i^l + Kn{h_i^l}^2))$\\
    \hline
    Image & Classical CNN & VGG-16& MNIST & 99.139 & -\\
    Image & Classical CNN & VGG-16& CIFAR-10 & 93.5 & -\\ \hline
    \end{tabular}
     \end{adjustbox}
\end{table*}
\subsection{Graph Convolutional Neural Network (GCNN)}
GCNNs can be broadly classified into two categories: Spectral-based and Spatial-based Models. Spectral-based models define graph convolutions by introducing filters from the perspective of graph signal processing~\cite{wu2019comprehensive}. Spatial-based models on the other hand define convolutions by information propagation. Spatial-based approaches have developed rapidly recently due to their efficiency, flexibility and generality. They are discussed in the following section.
\begin{figure}[b]
\centering
\includegraphics[scale=0.27]{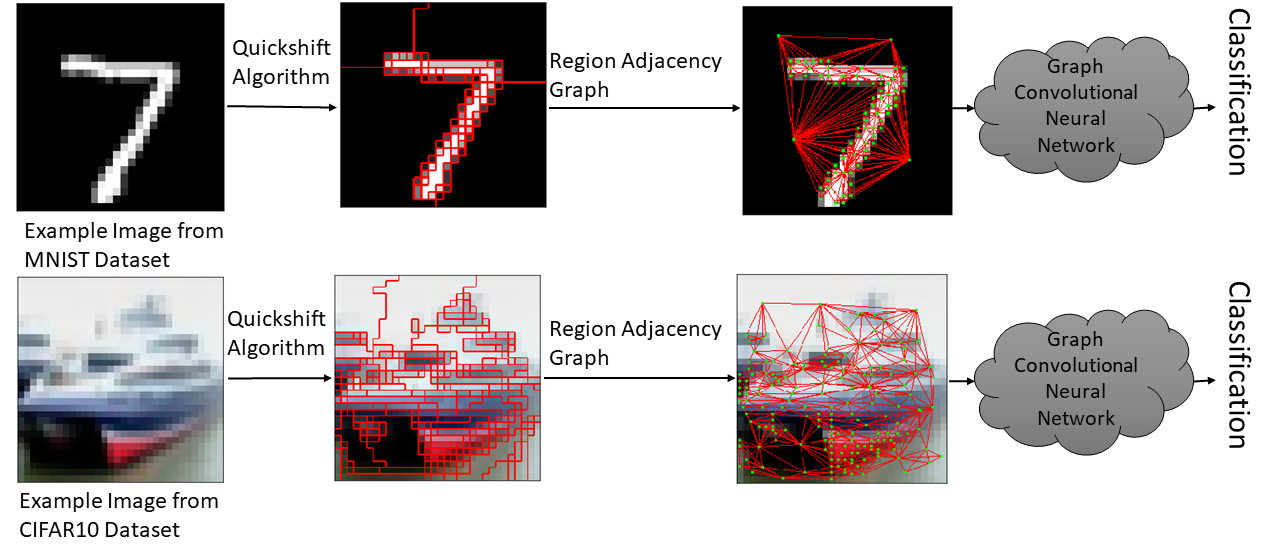}
\caption{Illustrative examples of our proposed classification approach}
\label{fig:illustrative}
\end{figure}
\subsubsection{Spectral Graph Convolutional Neural Networks}
Laplacian is the core operator in spectral graph theory. Laplacian matrix is the most natural form associated with a graph ($G$). Performing spectral conlvolution on graphs includes four steps as shown in Algorithm~\ref{algo:spectral}.
\begin{algorithm}
\KwIn{$x$, $D_G$, ${M}_G$}
\KwOut{$L_G$, $\hat{x}$, $y$}
Compute graph Laplacian: $L_G \stackrel{def}{=} D_G - {M}_G$ 
Weighted adjacency matrix: $W_G = {M}_G D_G^{-1}$ 
Laplacian matrix in normalized form: $L_G = I_n - D_G^{\frac{-1}{2}} W_G D_G^{\frac{-1}{2}}$
\\
Compute Fourier functions: $L_G = U \Lambda U^{T}$ \\
Compute Fourier transform: $\hat{x} = U^T x \in \R^n$ and
inverse Fourier transform: $x = U \hat{x}$ \\
Compute spectral convolution: 
\begin{align*}
    \begin{split}
        y & = h_\theta (L_G) x \\
        & = h_\theta (U \Lambda U^T) x \\
        & = U h_\theta(\Lambda_G) U^T x \\
    \end{split}
\end{align*}
\caption{Spectral Graph Convolution}
\label{algo:spectral}
\end{algorithm}

\begin{table}
    \caption{Quickshift Parameters}
    \label{parameters}
    \centering
    \begin{adjustbox}{width=0.5\textwidth}
    \begin{tabular}{cccccccc}\hline
    Dataset & Parameter & $\large\Bar{N}$ & $N_{min}$ & $N_{max}$ & $\large\Bar{deg}$ & $deg_{min}$ & $deg_{max}$ \\
    MNIST & $\epsilon=2, \alpha=1, S=2$ & 82.1 & 5 & 154 & 6.8 & 1 & 101 \\
    CIFAR-10 & $\epsilon=1, \alpha=1, S=5$ & 182 & 18 & 624 & 7.4 & 1 & 67 \\ \hline
    \end{tabular}
    \end{adjustbox}
\end{table}
However, there are several issues with this convolutional structure. First, the eigenvector matrix $U$
requires the explicit computation of the eigenvalue decomposition of the graph Laplacian matrix, and hence suffers from the $\mathcal{O}(n^3)$ time complexity which is impractical for large-scale graphs. Second, though the eigenvectors can be precomputed, the time complexity of Algorithm~\ref{algo:spectral} is still $\mathcal{O}(n^3)$. Third, there are $\mathcal{O}(n)$ parameters to be learned in each layer. Besides, these non-parametric filters are not localized in the vertex domain.

To address these limitations, the ChebNet~\cite{defferrard2016convolutional} proposed the use of K-polynomial filters in the convolutional layers for localization. To compute fast spectral convolutions in $\mathcal{O}(n)$ time, we have to define compact kernels~\cite{defferrard2016convolutional}. Let a non-parametric filter, i.e., a filter whose parameters are all free be defined as
\begin{equation}
    h_\theta(\Lambda) = diag(\theta)
\end{equation}
where parameter $\theta \in \R^n$ is a vector of Fourier coefficients. There are however two limitations with non-parametric filters: (1) They are not localized in space and (2) Their learning complexity is in $\mathcal{O}(n)$, the dimensionality of the data.
These issues can be overcome with the use of a polynomial filter~\cite{defferrard2016convolutional}:
\begin{equation}\label{eqpfilter}
    h_\theta (\Lambda) = \sum_{k=0}^{K-1} \theta_k \Lambda^k, 
\end{equation}
where the parameters $\theta \in \R^K$ is a vector of polynomial coefficients.
By \cite{defferrard2016convolutional}, \eqref{eqpfilter} can be expressed as Chebyshev polynomial
\begin{equation}
    h_\theta (\Lambda) = \sum_k \theta_k T_k(\hat{\Lambda})
\end{equation}
of order $K-1$, where the parameter $\theta \in \R^K$ is a vector of Chebyshev coefficients and $T_k(\hat{\Lambda}) \in \R^{n \times n}$ is the Chebyshev polynomial of order $K$ evaluated at $\hat{\Lambda}= \frac{2 \Lambda}{\lambda_{max}-I_n}$, a diagonal matrix of scaled eigen values that lie in $[-1, 1]$.

\begin{algorithm}
\KwIn{Superpixels/Node embeddings}
\KwOut{Prediction score}
Each neighbor sends a message: ${\bold m}_{w,v}^{(l)} = MESSAGE({\bold h}_v^{l-1},{\bold h}_w^{l-1})$\\
Messages are aggregated across all neighbors: ${\bold a}_v^l = AGGREGATE({{\bold m}_{w,v}^{(l)}: w \in \mathcal{N}(v)})$ \\
Neighbor information is used to update node representation: ${\bold h}_v^l = UPDATE({\bold h}_v^{l-1}, {\bold a}_v^{l})$ \\
Prediction based on global readout of node embeddings:
${\bold \phi}(G) = MLP(\sum_{v \in V} {\bold h}_v^{(L)})$
\caption{Spatial Graph Convolution}
\label{algo:spatial}
\end{algorithm}

\subsubsection{Spatial Graph Convolutional Neural Network}
In this subsection, we focus on the spatial graph convolutions that propagate and aggregate the node representations from neighboring nodes in the vertex domain~\cite{wu2019comprehensive,fey2019fast} as shown in Algorithm~\ref{algo:spatial}. The most popular implementation of spatial convolution filtering is given as
\begin{equation}
 \small   {\bf h}_v^{l} = \sigma({\bf W}_1 {\bf h}_v^{(l-1)} + {\bf W}_2 \sum_{w \in \mathcal{N}(v)} C_{v, w}{\bf h}_w{(l-1)} )
\end{equation}
Where $W_1, W_2$ are trainable parameters and $C_{v,w}$ is Normalization coefficient and structure dependent, i.e., $C_{v,w}=|\mathcal{N}(v)|^{-1}$.
\section{Experimental Evaluation}
We evaluate three models: ChebNet~\cite{defferrard2016convolutional}, EGCNN~\cite{fey2019fast} and GNN baseline on image based graph classification dataset: MNIST and CIFAR-10, in comparison with classical CNN model: VGG-16. ChebNet is spectral-based model while GNN baseline and EGCNN are spatial-based models.
For each dataset there is a set of graphs with different number of nodes (superpixels) and each graph $G$ has a single categorical label that is to be predicted. Dataset details, parameters and metrics are discussed in following subsections.
\subsection{Datasets}
The first dataset is MNIST (Modified National Institute of Standards and Technology database). It is a large database of handwritten digits from $0$ to $9$. It consists of $55,000$ training images, $5000$ validation images and $10,000$ test images. The images are uniform on the scale $28\times 28$ pixels and have only one color channel with grey values. The other dataset is CIFAR-10 (Canadian Institute for Advanced Research dataset). It is a large dataset of $60,000$ color images, each of which is exactly one of $10$ classes: airplane, car, bird, cat, deer, dog, frog, horse, ship and truck. It consists of $45,000$ training images, $5000$ validation images and $10,000$ test images. The images are uniform size of scale $32 \times 32$ pixels and have three color channels.
As shown in Fig.~\ref{fig:illustrative}, $2$ preprocessing steps are performed on both of these datasets. The first step is to generate superpixels using Quickshift~\cite{vedaldi2008quick} (see section~\ref{qs}) and second step is graph generation process from superpixels (see section~\ref{rag}). Using the Quickshift algorithm, data reduction is approximately $30\%$ of input data i.e. $2150$ characteristics of image graph as compared to the number of features of the original image ($32 \times 32 \times 3 = 3072$).
\subsection{Parameters and Metrics}
Quickshift parameters for MNIST and CIFAR-10 datasets are shown in table~\ref{parameters}. We use Softmax Regression~\ref{softm} for multidimensional classification problem, that is the probability distribution of classes $Y$ for the occurrence of input $\bold x$. The exponential function in equation~\ref{softm} highlighted the highest values and weakened the significant values which are below the maximum.
\begin{equation}\label{softm}
 softmax(\bold y) = \frac{exp({\bold y})}{\sum_i^Y exp({\bold y}_i)}
\end{equation}
The cross-entropy loss is defined over an input set $X$ as
\begin{equation}
H(\mathcal{X}) = \frac{1}{|\mathcal{X}|} \sum_{x \in \mathcal{X}}( \hat{y}({\bold x}), log(y({\bold x})))
\end{equation}
To evaluate the results of a classification, the accuracy over an input set $X$ as
\begin{equation}
accuracy(\mathcal{X}) = \frac{1}{|\mathcal{X}|} \sum_{x \in \mathcal{X}} max(-f + 1, 0)
\end{equation}
where $f = |argmax(y({\bold x}))- argmax(\hat{y}({\bold x}))|$ and $argmax(y) = n$ if and only if $y_n >= y_m$ for $1<=m<=Y$. This describes the proportion of correctly classified input data from the respective total amount of training, validation and test data $X$ and is therefore a suitable mean of determining the quality of classification network.

\subsection{Quantitative Evaluation}
As we can see in table~\ref{tab:eval}, overall spectral-based GNN: ChebNet~\cite{defferrard2016convolutional} perform better than the Embedded GCN model and classical model: VGG-16. Although the GCN make use of pixel or superpixel based RAG and classical CNN utilizes original images for classification but the accuracy of ChebNet~\cite{defferrard2016convolutional} is better than all other models with lesser compute cost.
\section{Conclusion}
In this survey paper, we thoroughly reviewed and provided several ways for image classification utilizing graph neural networks. Our purposed technique suggests to convert image into superpixels using Quickshift algorithm and then use superpixles to generate region adjacency graph. For classification task, this graph is passed through different GNN models. We also analyzed the spatial and spectral convolutions filtering techniques for classification. Spectral-based models perform better than spatial-based models and classical CNN with lesser compute cost. 
\footnotesize 
\bibliographystyle{IEEEtran}
\bibliography{mainbib}


\end{document}